\documentclass[10pt,twocolumn,letterpaper]{article}

\usepackage{cvpr}
\usepackage{times}
\usepackage{epsfig}
\usepackage{graphicx}
\usepackage{amsmath}
\usepackage{amssymb}

\usepackage{caption}
\usepackage{color}
\usepackage{subfigure}
\usepackage{lipsum}
\usepackage{epstopdf}
\usepackage{array}

\usepackage{bbding}

\usepackage[american]{babel} 
\usepackage{cite}

\usepackage[pagebackref=true,breaklinks=true,letterpaper=true,colorlinks,urlcolor=black,bookmarks=false]{hyperref}

\cvprfinalcopy 

\def\cvprPaperID{1439} 

\ifcvprfinal\fi
\begin{document}

\newcommand*{\affaddr}[1]{#1} 
\newcommand*{\affmark}[1][*]{\textsuperscript{#1}}
\newcommand*{\email}[1]{\texttt{#1}}
\newcommand\blfootnote[1]{%
\begingroup
\renewcommand\thefootnote{}\footnote{#1}%
\addtocounter{footnote}{-1}%
\endgroup
}
\newcommand{\tabincell}[2]{\begin{tabular}{@{}#1@{}}#2\end{tabular}}
\title{STGAN: A Unified Selective Transfer Network \\for Arbitrary Image Attribute Editing}

\author{%
Ming Liu\affmark[1$^\star$],
Yukang Ding\affmark[2],
Min Xia\affmark[1],
Xiao Liu\affmark[2],
Errui Ding\affmark[2],
Wangmeng Zuo$^{(}$\Envelope$^{)}$\affmark[1,3],
Shilei Wen\affmark[2]\\
\affaddr{\small \affmark[1]Harbin Institute of Technology},
\affaddr{\small \affmark[2]Department of Computer Vision Technology (VIS), Baidu Inc.},
\affaddr{\small \affmark[3]Peng Cheng Laboratory, Shenzhen}\\
\email{\small csmliu@outlook.com,}
\email{\small csmxia@gmail.com,}
\email{\small wmzuo@hit.edu.cn,}\\
\texttt{\small \{\href{mailto:dingyukang@baidu.com}{dingyukang}, \href{mailto:liuxiao12@baidu.com}{liuxiao12},
                 \href{mailto:dingerrui@baidu.com}{dingerrui}, \href{mailto:wenshilei@baidu.com}{wenshilei}\}@baidu.com}
}


\twocolumn[{%
\renewcommand\twocolumn[1][]{#1}%
\maketitle
\begin{center}
    \centering
    \vspace{-1.8em}
    \includegraphics[width=.95\linewidth]{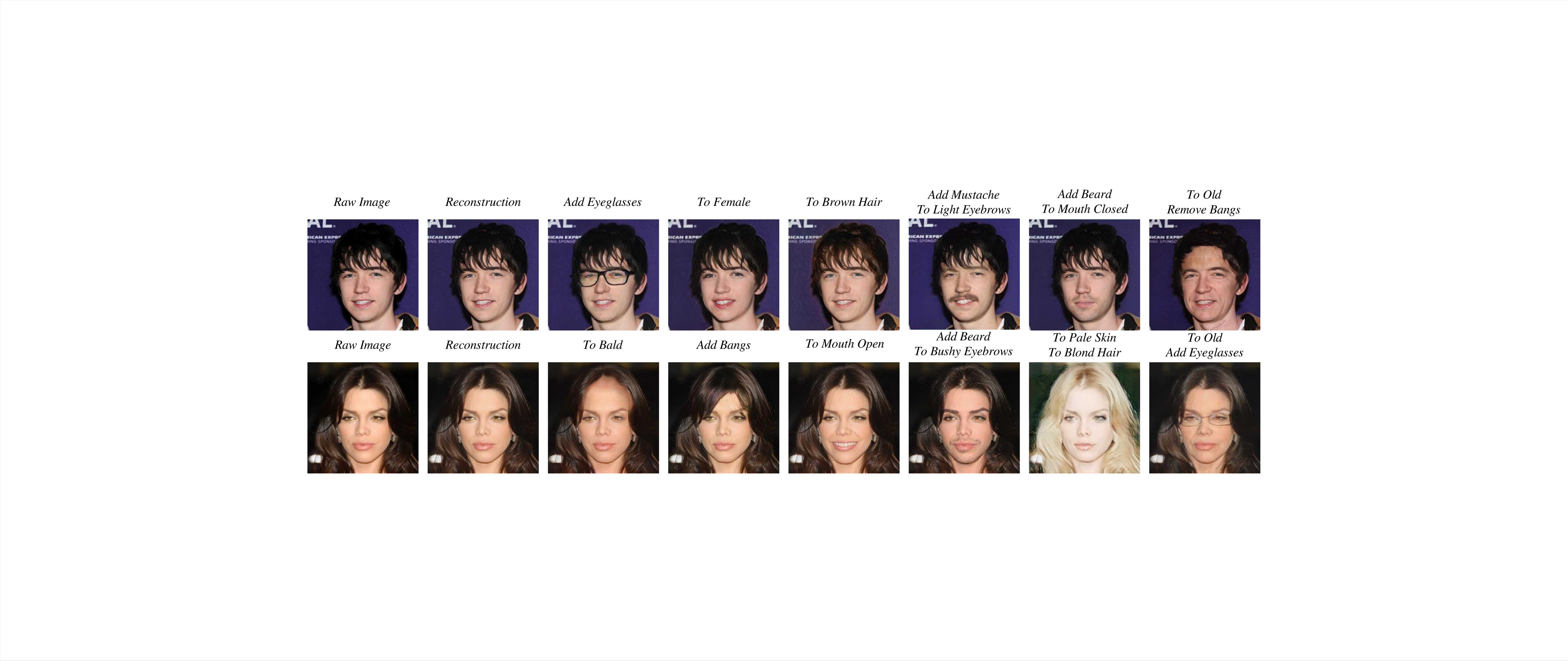}\\
    \vspace{-.75em}
    \captionsetup{font=small}
    \captionof{figure}{High resolution ($384\times384$) results of STGAN for facial attribute editing, and more results are given in the suppl.}%
    \label{fig:HD_results}
\end{center}%
}]

\begin{abstract}
\vspace{-.5em}
   Arbitrary attribute editing generally can be tackled by incorporating encoder-decoder and generative adversarial networks.
   However, the bottleneck layer in encoder-decoder usually gives rise to blurry and low quality editing result.
   And adding skip connections improves image quality at the cost of weakened attribute manipulation ability.
   Moreover, existing methods exploit target attribute vector to guide the flexible translation to desired target domain.
   In this work, we suggest to address these issues from selective transfer perspective.
   Considering that specific editing task is certainly only related to the changed attributes instead of all target attributes, our model selectively takes the difference between target and source attribute vectors as input.
   Furthermore, selective transfer units are incorporated with encoder-decoder to adaptively select and modify encoder feature for enhanced attribute editing.
   Experiments show that our method (\ie, STGAN) simultaneously improves attribute manipulation accuracy as well as perception quality, and performs favorably against state-of-the-arts in arbitrary facial attribute editing and season translation.

\end{abstract}
\vspace{-1.5em}
\blfootnote{$^\star$Work done during an internship at Baidu.}

\section{Introduction}\label{section:Introduction}
\vspace{-.1em}
Image attribute editing, aiming at manipulating an image to possess desired attributes, is an interesting but challenging problem with many real-world vision applications.
On one hand, it is impracticable to collect paired images with and without desirable attributes (\eg, \emph{female} and \emph{male} face images of the same person).
Thus, unsupervised generative learning models, \eg, generative adversarial networks (GANs)~\cite{goodfellow2014generative}, have attracted upsurging attention in attribute editing.
On the other hand, arbitrary attribute editing actually is a multi-domain image-to-image translation task.
Learning single translation model for each specific attribute editing task may achieve limited success~\cite{li2016DIAT, zhu2017unpaired, shen2017learning}.
But it is ineffective in exploiting the entire training data, and the learned models grows exponentially along with the number of attributes.
To handle this issue, several arbitrary attribute editing approaches~\cite{perarnau2016invertible, he2017arbitrary, Choi_2018_CVPR} have been developed, which usually (i)~use encoder-decoder architecture, and (ii) take both source image and target attribute vector as input.

Albeit their extensive deployment, encoder-decoder networks remain insufficient for high quality attribute editing.
Attribute can be of either local, global, or abstract characteristic of the image.
In order to properly manipulate image attribute, spatial pooling or downsampling generally are required to obtain high-level abstraction of image content and attributes.
For example, auto-encoder architecture is adopted in~\cite{perarnau2016invertible,lample2017fader,he2017arbitrary}, and shallow encoder-decoder with residual blocks is used in~\cite{zhu2017unpaired,Choi_2018_CVPR}.
However, the introduction of bottleneck layer, \ie, the innermost feature map with minimal spatial size, gives rise to blurry and low quality editing result.
As a remedy, some researchers suggest to add one~\cite{he2017arbitrary} or multiple~\cite{isola2017image} skip connections between encoder and decoder layers.
Unfortunately, as further shown in Sec. \ref{subsection:skip_conn}, the deployment of skip connections improves image quality of editing result but is harmful to attribute manipulation ability of learned model.
Another possible solution is to employ spatial attention network to allow attribute-specific region editing~\cite{zhang2018generative}, which, however, is effective only for local attributes and not designed for arbitrary attribute editing.

\begin{figure}[t]
    \centering
    \captionsetup[subfigure]{labelformat=simple, listofformat=subsimple}
    \begin{minipage}{0.23\linewidth}
        \includegraphics[width=\linewidth]{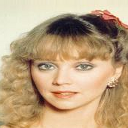}
        \includegraphics[width=\linewidth]{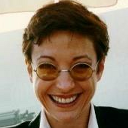}
        \vspace{-2em}
        \caption*{\scriptsize Input}
    \end{minipage}
    \begin{minipage}{0.23\linewidth}
        \includegraphics[width=\linewidth]{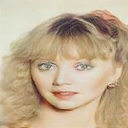}
        \includegraphics[width=\linewidth]{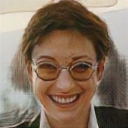}
        \vspace{-2em}
        \caption*{\scriptsize AttGAN}
    \end{minipage}
    \begin{minipage}{0.23\linewidth}
        \includegraphics[width=\linewidth]{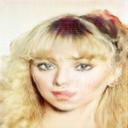}
        \includegraphics[width=\linewidth]{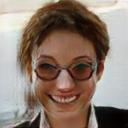}
        \vspace{-2em}
        \caption*{\scriptsize StarGAN}
    \end{minipage}
    \begin{minipage}{0.23\linewidth}
        \includegraphics[width=\linewidth]{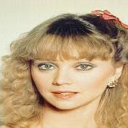}
        \includegraphics[width=\linewidth]{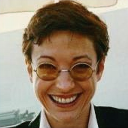}
        \vspace{-2em}
        \caption*{\scriptsize STGAN}
    \end{minipage}
    \setlength{\abovecaptionskip}{0pt}
    \setlength{\belowcaptionskip}{-5pt}
    \captionsetup{font=small}
    \caption{Reconstruction results of AttGAN~\cite{he2017arbitrary}, StarGAN~\cite{Choi_2018_CVPR} and our STGAN.}
    \label{fig:face_rec_image}
    \vspace{-1em}
\end{figure}

Moreover, most existing methods exploit both source image and target attribute vector for arbitrary attribute editing.
In particular, the encoders in~\cite{lample2017fader,he2017arbitrary} only take source image as input to produce latent code, and then the decoders utilize both latent code and target attribute vector to generate editing result.
In contrast, StarGAN~\cite{Choi_2018_CVPR} directly takes source image and target attribute vector as input.
%
However, for arbitrary attribute editing only the attributes to be changed are required, taking full target attribute vector as input may even have adverse effect on editing result.
As shown in Fig.~\ref{fig:face_rec_image}, although all attributes keep unchanged, unwanted changes and visual degradation can be observed in the results by AttGAN~\cite{he2017arbitrary} and StarGAN~\cite{Choi_2018_CVPR}, mainly ascribing to the limitation of encoder-decoder and the use of target attribute vector as input.

%
To address the above issues, this work investigates arbitrary attribute editing from selective
transfer perspective and presents a STGAN model.
In terms of \emph{selective}, our STGAN is suggested to (i) only consider the attributes to be changed, and (ii) selectively concatenate encoder feature in editing attribute irrelevant regions with decoder feature.
In terms of \emph{transfer}, our STGAN is expected to adaptively modify encoder feature to match the requirement of varying editing task, thereby providing a unified model for handling both local and global attributes.

To this end, instead of full target attribute vector, our STGAN takes the difference between target and source attribute vectors as input to encoder-decoder.
Subsequently, selective transfer units (STUs) are proposed to adaptively select and modify encoder feature, which is further concatenated with decoder feature for enhancing both image quality and attribute manipulation ability.
In particular, STU is added to each pair of encoder and decoder layers, and takes both encoder feature, inner state, and difference attribute vector into consideration for exploiting cross-layer consistency and task specificity.
From Figs. \ref{fig:HD_results} and \ref{fig:face_rec_image}, our STGAN can generate high quality and photo-realistic results for arbitrary attribute editing, and obtain near-ideal reconstruction when the target and source attributes are the same.
%
%
To sum up, the contribution of this work involves:
\begin{itemize}
  \item Instead of all target attributes, difference attribute vector is taken as input to enhance the flexible translation of attributes and ease the training procedure.
  \vspace{-2mm}
  \item Selective transfer units are presented and incorporated with encoder-decoder for simultaneously improving attribute manipulation ability and image quality.
  \vspace{-2mm}
  \item Experimental results show that our STGAN performs favorably against state-of-the-arts in arbitrary facial attribute editing and season translation.
\end{itemize}

\section{Related Work}\label{section:RelatedWork}
\vspace{-0.2em}
\noindent\textbf{Encoder-Decoder Architecture.}
In their pioneer work~\cite{hinton1994autoencoders}, Hinton and Zemel proposed an autoencoder network,       
which consists of an encoder to map the input into \emph{latent code} and a decoder
to recover from the \emph{latent code}.
Subsequently, denoising autoencoders \cite{Vincent2008Extracting} are presented to learn representation robust to partial corruption.
%
%
Kingma and Welling~\cite{kingma2013auto} suggested a Variational Autoencoder (VAE),    
which validates the feasibility of encoder-decoder architecture to generate unseen images.
%
Recent studies show that skip connections~\cite{ronneberger2015u,isola2017image} between encoder and decoder layers usually benefit the training stability and visual quality of generated images.
However, as discussed in Sec.~\ref{subsection:skip_conn}, skip connections actually improves image quality at the cost of weakened attribute manipulation ability, and should be carefully used in arbitrary attribute editing.
\vspace{0.5em}

\noindent\textbf{Generative Adversarial Networks.}
GAN~\cite{goodfellow2014generative, 
radford2015unsupervised} is originally proposed to generate images from random noise, and generally consists of a generator and a discriminator which are trained in an adversarial manner and suffer from the mode collapse problem.
Recently, enormous efforts have been devoted to improving the stability of learning. 
In~\cite{arjovsky2017wasserstein,gulrajani2017improved}, Wasserstein-1 distance and gradient penalty are suggested to improve stability of the optimization process.
In~\cite{larsen2016autoencoding}, the VAE decoder and GAN generator are collapsed into one model and optimized by both reconstruction and adversarial loss.
Conditional GAN (cGAN)~\cite{mirza2014conditional,isola2017image} takes conditional variable as input to the generator and discriminator to generate image with desired properties.
As a result, GAN has become one of the most prominent models for versatile image generation~\cite{goodfellow2014generative, radford2015unsupervised}, translation~\cite{isola2017image,zhu2017unpaired}, restoration~\cite{ledig2017photo,Li_2018_ECCV} and editing~\cite{pathak2016context} tasks.
\vspace{0.5em}

\noindent\textbf{Image-to-Image Translation.}
Image-to-image translation aims at learning cross-domain mapping in supervised or unsupervised settings.
Isola \etal~\cite{isola2017image} presented a unified pix2pix framework for learning image-to-image translation from paired data.
Improved network architectures, \eg, cascaded refinement networks~\cite{chen2017photographic} and pix2pixHD~\cite{Wang_2018_CVPR}, are then developed to improve the visual quality of synthesized images.
As for unpaired image-to-image translation, additional constraints, \eg, cycle consistency~\cite{zhu2017unpaired} and shared latent space~\cite{liu2017unsupervised}, are suggested to alleviate the inherent ill-posedness of the task.
Nonetheless, arbitrary attribute editing actually is a multi-domain image-to-image translation problem, and cannot be solved with scalability by aforementioned methods.
To address this issue, \cite{anoosheh2018ComboGAN} and \cite{hui2017unsupervised} decouple generators by learning domain-specific encoders/decoders with shared latent space, but are still limited in scaling to change multiple attributes of an image.
\vspace{0.5em}

\noindent\textbf{Facial Attribute Editing.}
Facial attribute editing is an interesting multi-domain image-to-image translation problem and has received considerable recent attention.
While several methods have been proposed to learn single translation model for each specific attribute
editing task~\cite{li2016DIAT,shen2017learning,Chen_2018_CVPR,zhang2018generative}, they suffer from the limitation of image-to-image translation and cannot well scale to arbitrary attribute editing.
Therefore, researchers resort to learning a single model for arbitrary attribute editing.
IcGAN~\cite{perarnau2016invertible} adopts an encoder to generate latent code of an image, and a cGAN to decode latent code conditioned on target attributes.
However, IcGAN first trains the cGAN model followed by the encoders, greatly restricting its reconstruction ability.
Lample \etal~\cite{lample2017fader} trained the FaderNet in an end-to-end manner by imposing
adversarial constraint to enforce the independence between latent code and attributes.
ModularGAN~\cite{Zhao_2018_ECCV} presents a feasible solution to connect specific attribute editing to arbitrary attribute editing, but its computation time gradually increases along with the number of attributes to be changed.
StarGAN~\cite{Choi_2018_CVPR} and AttGAN~\cite{he2017arbitrary} elaborately tackle arbitrary attribute editing by taking target attribute vector as input to the transform model.
In this work, we analyze the limitation of StarGAN~\cite{Choi_2018_CVPR} and AttGAN~\cite{he2017arbitrary}, and further develop a STGAN for simultaneously enhancing the
attribute manipulation ability and image quality.

\begin{figure}
    \centering
    \captionsetup[subfigure]{labelformat=simple, listofformat=subsimple}
    \begin{minipage}{0.18\linewidth}
        \includegraphics[width=\linewidth]{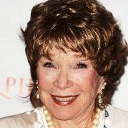}
        \vspace{-2.2em}
        \caption*{\scriptsize Input}
    \end{minipage}
    \begin{minipage}{0.18\linewidth}
        \includegraphics[width=\linewidth]{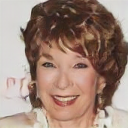}
        \vspace{-2.2em}
        \caption*{\scriptsize AttGAN-ED}
    \end{minipage}
    \begin{minipage}{0.18\linewidth}
        \includegraphics[width=\linewidth]{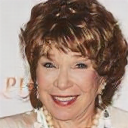}
        \vspace{-2.2em}
        \caption*{\scriptsize AttGAN}
    \end{minipage}
    \begin{minipage}{0.18\linewidth}
        \includegraphics[width=\linewidth]{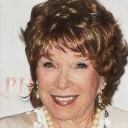}
        \vspace{-2.2em}
        \caption*{\scriptsize AttGAN-2s}
    \end{minipage}
    \begin{minipage}{0.18\linewidth}
        \includegraphics[width=\linewidth]{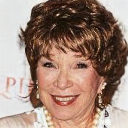}
        \vspace{-2.2em}
        \caption*{\scriptsize AttGAN-UNet}
    \end{minipage}
    \setlength{\abovecaptionskip}{5pt}
    \setlength{\belowcaptionskip}{-5pt}
    \captionsetup{font=small}
    \caption{Results of AttGAN~\cite{he2017arbitrary} variants for
    reconstructing input image. Please zoom in for better observation.}
    \label{fig:attgan_img_comp}
    \vspace{-1em}
\end{figure}

\begin{table}
    \centering
    \scalebox{0.7}{
    \begin{tabular}{lcccc}
        \hline
        {Method} & {AttGAN-ED} & {AttGAN} &
                            {AttGAN-2s} & {AttGAN-UNet} \\ \hline
        {PSNR/SSIM} & 22.68/0.758 & 24.07/0.841
                            & 26.13/0.897 & \textbf{29.66}/\textbf{0.929} \\ \hline%
    \end{tabular} }
    \setlength{\abovecaptionskip}{3pt}
    \setlength{\belowcaptionskip}{0pt}
    \captionsetup{font=small}
    \caption{Reconstruction evaluation of {\small AttGAN~\cite{he2017arbitrary}} variants.}
    \label{tab:attgan_rec_comp}
    \vspace{-1em}
\end{table}

\begin{figure}
    \centering
    \includegraphics[width=\linewidth]{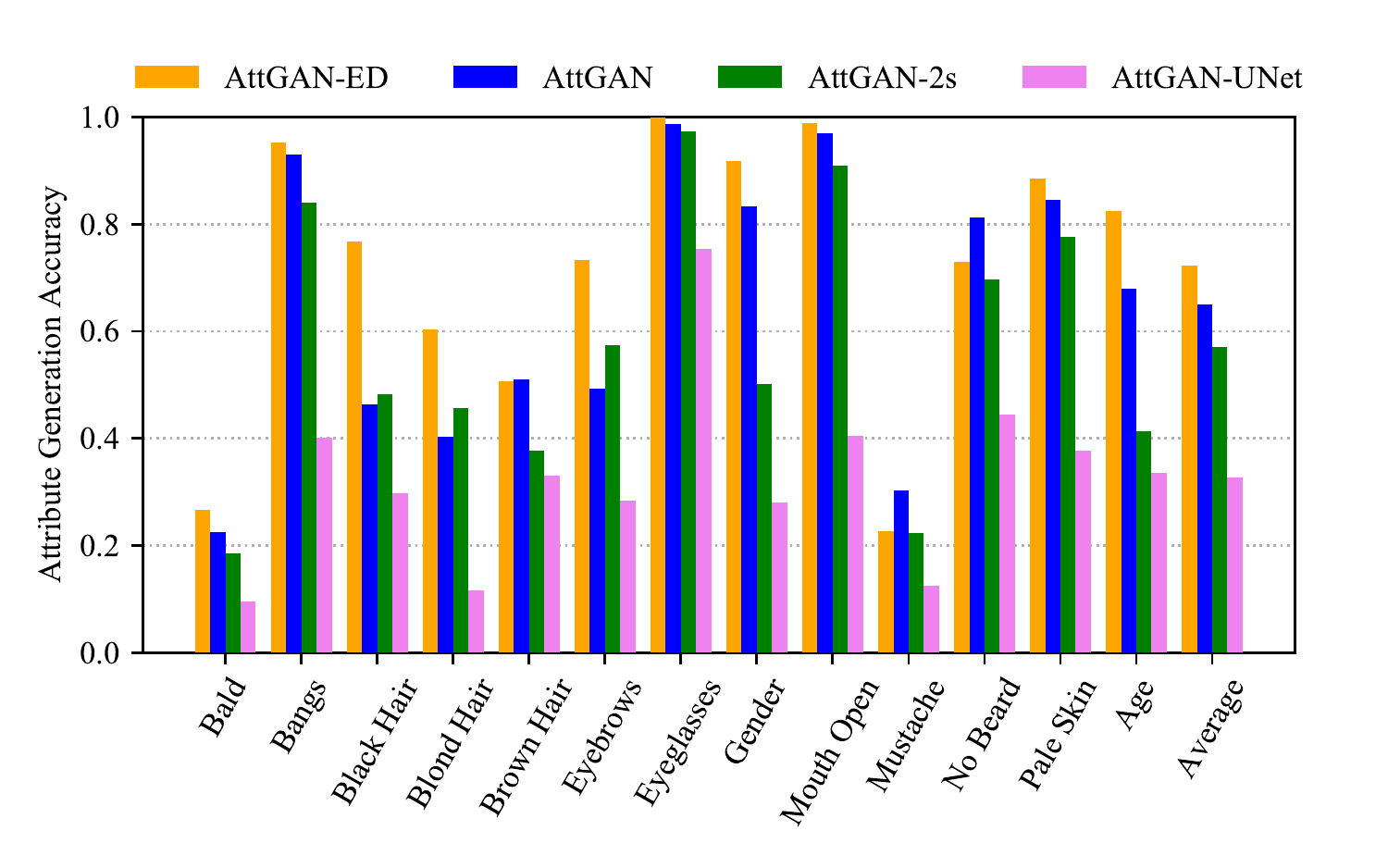}
    \setlength{\abovecaptionskip}{-10pt}
    \setlength{\belowcaptionskip}{-5pt}
    \captionsetup{font=small}
    \caption{Attribute generation accuracy of AttGAN~\cite{he2017arbitrary} variants.}
    \label{fig:attgan_acc_comp}
    \vspace{-1em}
\end{figure}

\vspace{-0.3em}
\section{Proposed Method}\label{section:Method}
\vspace{-0.2em}
This section presents our proposed STGAN for arbitrary attribute editing.
To begin with, we use AttGAN as an example to analyze the limitation of skip connections.
%
%
Then, we formulate STGAN by taking difference attribute vector as input and incorporating selective transfer units into encoder-decoder structure.
Finally, network architecture (see Fig.~\ref{fig:structure}) and model objective of STGAN are provided.

\vspace{-0.2em}
\subsection{Limitation of Skip Connections in AttGAN}\label{subsection:skip_conn}
\vspace{-0.2em}
StarGAN~\cite{Choi_2018_CVPR} and AttGAN~\cite{he2017arbitrary} adopt encoder-decoder structure, where spatial pooling or downsampling are essential to obtain high level abstract representation for attribute manipulation.
Unfortunately, downsampling irreversibly diminishes spatial resolution and fine details of feature map, which cannot be completely recovered by transposed convolutions and the results are prone to blurring or missing details.
To enhance image quality of editing result, AttGAN~\cite{he2017arbitrary} applies one skip connection between encoder and decoder, but we will show that it is still limited.

To analyze the effect and limitation of skip connections, we test four variants of AttGAN on the test set: (i) AttGAN w/o skip connection (AttGAN-ED), (ii) AttGAN model released by He~\etal~\cite{he2017arbitrary} with one skip connection (AttGAN), (iii) AttGAN with two skip connections (AttGAN-2s), and (iv) AttGAN with all symmetric skip connections~\cite{ronneberger2015u} (AttGAN-UNet).
Table~\ref{tab:attgan_rec_comp} lists the PSNR/SSIM results of reconstruction by keeping target attribute vector the same as the source one, and Fig.~\ref{fig:attgan_img_comp} shows the reconstruction results of an image.
It can be seen that adding skip connections does benefit the reconstruction of fine details, and better result can be obtained with the increase of skip connections.
By setting target attribute vector different from source one, Fig.~\ref{fig:attgan_acc_comp} further assesses the facial attribute generation accuracy via a facial attribute
classification model\footnote{We train the model on CelebA~\cite{liu2015faceattributes}
dataset which can achieve $94.5\%$ mean accuracy on the 13 attributes we use.}.
While adding one skip connection, \ie, AttGAN, only slightly decreases generation accuracy for most attributes, notable degradation can be observed by adding multiple skip connections.
Thus, the deployment of skip connections improves reconstruction image quality at the cost of weakened attribute manipulation ability, mainly attributing to that skip connection directly concatenates encoder and decoder features.
To circumvent this dilemma, we present our STGAN to employ selective transfer units to adaptively transform encoder features guided by attributes to be changed.
%

\subsection{Taking Difference Attribute Vector as Input}
Both StarGAN~\cite{Choi_2018_CVPR} and AttGAN~\cite{he2017arbitrary} take target attribution vector $\mathbf{att}_t$ and source image $\mathbf{x}$ as input to the generator.
Actually, the use of full target attribution vector is redundant and may be harmful to editing result.
In Fig.~\ref{fig:face_rec_image}, the target attribution vector $\mathbf{att}_t$ is exactly the same as on the source one $\mathbf{att}_s$, but StarGAN~\cite{Choi_2018_CVPR} and AttGAN~\cite{he2017arbitrary} may manipulate some unchanged attributes by mistake.
From Fig.~\ref{fig:face_rec_image}, after editing the face image with blond hair becomes more blond.
Moreover, they even incorrectly adjust hair length of a source image with the attribute \emph{female}.

For arbitrary image attribute editing, instead of full target attribute vector, only the attributes to be changed should be considered to preserve more information of source image.
So we define the difference attribute vector as the difference between target and source attribute vectors,
    \vspace{-2mm}
    \begin{equation}\label{eqn:diff_label}
        \mathbf{att}_\mathit{diff} = \mathbf{att}_\mathit{t} - \mathbf{att}_\mathit{s}.
    \end{equation}
%
Taking $\mathbf{att}_\mathit{diff}$ as input can bring several distinctive merits.
First, the attributes to be changed are only a small set of attribute vector, and the use of $\mathbf{att}_\mathit{diff}$ usually makes the model easier to train.
Second, in comparison to $\mathbf{att}_\mathit{t}$, $\mathbf{att}_\mathit{diff}$ can provide more valuable information for guiding image attribute editing, including whether an attribute is required to edit or not, toward what direction an attribute should be changed.
The information can then be utilized to design proper model to transform and concatenate encoder feature with decoder feature, and improve image reconstruction quality without sacrifice of attribute manipulation accuracy.
Finally, in practice $\mathbf{att}_\mathit{diff}$ actually is more convenient to be provided by user.
When taking $\mathbf{att}_\mathit{t}$ as input, the user is required to either manually supply all target attributes, or modify source attributes provided by some attribute prediction method.

\subsection{Selective Transfer Units}\label{subsection:STU}
\begin{figure*}[ht]
    \centering
    \scalebox{0.95}{\includegraphics[width=\linewidth]{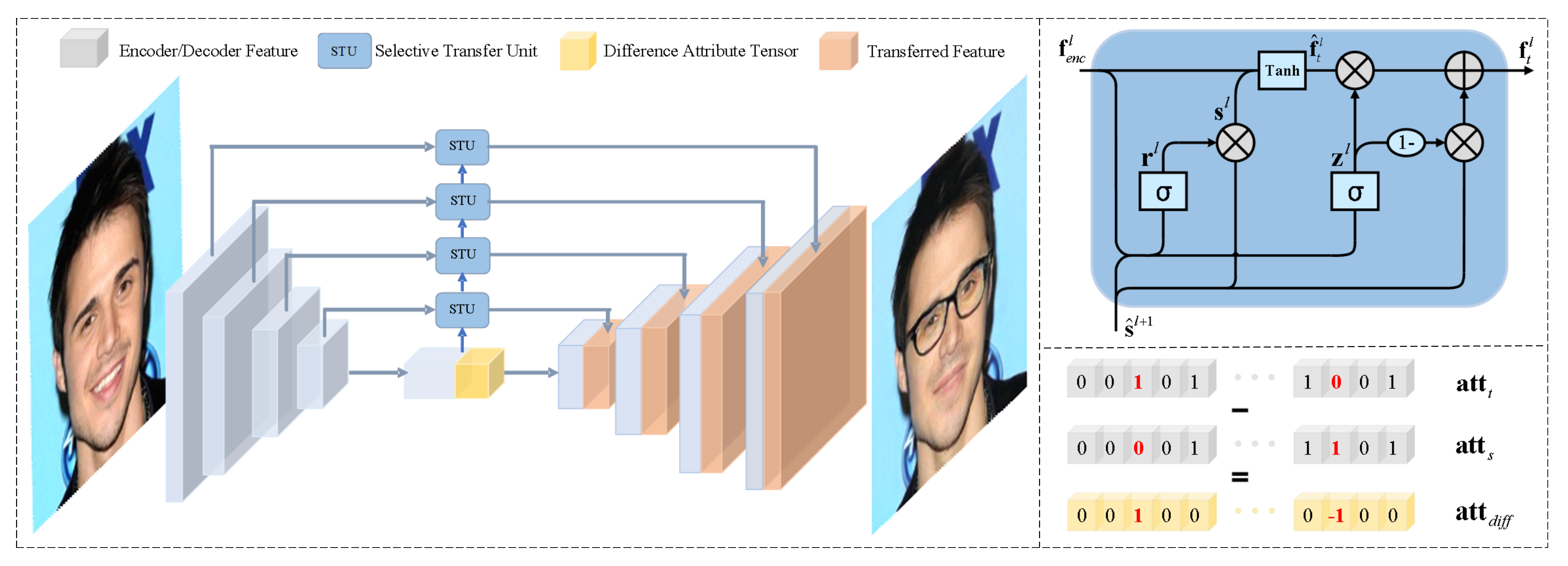}}
    \captionsetup{font=small}
    \caption{The overall structure of STGAN. On the left is the generator. The top-right figure shows detailed STU structure, and all variables marked in this figure share same dimension (\eg, $64\times64$).
             The difference attribute vector of adding \emph{Eyeglasses} and removing \emph{Mouth Open} attributes is shown on the bottom-right.}
    \setlength{\abovecaptionskip}{-20pt}
    \setlength{\belowcaptionskip}{-5pt}
    \label{fig:structure}
    \vspace{-1.5em}
\end{figure*}

%
Fig.~\ref{fig:structure} shows the overall architecture of our STGAN.
Instead of directly concatenating encoder with decoder features via skip connection, we present selective transfer unit (STU) to selectively transform encoder feature, making it compatible and complementary to decoder feature.
Naturally, the transform is required to be adaptive to the changed attributes, and be consistent among different encoder layers.
Thus, we modify the structure of GRU~\cite{cho2014learning, chung2014empirical} to build STUs for passing information from inner layers to outer layers.

Without loss of generality, we use the $\mathit{l}$-th encoder layer as an example.
Denote by $\mathbf{f}_\mathit{enc}^\mathit{l}$ the encoder feature of the $\mathit{l}$-th layer, and $\mathbf{s}^{\mathit{l}+1}$ the hidden state from the $\mathit{l}+1$-th layer.
For convenience, the difference attribute vector $\mathbf{att}_\mathit{diff}$ is stretched to have the same spatial size of $\mathbf{s}^{\mathit{l}+1}$.
Different from sequence modeling, feature maps across layers are of different spatial size.
So we first use transposed convolution to upsample hidden state $\mathbf{s}^{\mathit{l}+1}$,
    \vspace{-1mm}
    \begin{equation}\label{eqn:state}
        \hat{\mathbf{s}}^{\mathit{l}+1} = \mathbf{W}_t \ast_T [\mathbf{s}^{\mathit{l}+1},
            \mathbf{att}_\mathit{diff}],
    \end{equation}
where $[\cdot, \cdot]$ denotes the concatenation operation, and $\ast_T$ denotes transposed convolution.
Then, STU adopts the mathematical model of GRU to update the hidden state $\mathbf{s}^{l}$ and transformed encoder feature $\mathbf{f}_{t}^{l}$,
    \vspace{-1mm}%
    \begin{equation}\label{eqn:reset_gate}
        \mathbf{r}^\mathit{l} = \sigma(\mathbf{W}_\mathit{r} \ast [\mathbf{f}_\mathit{enc}^\mathit{l}, \hat{\mathbf{s}}^{\mathit{l}+1}]),
    \end{equation}
    \vspace{-2mm}%
    \begin{equation}\label{eqn:update_gate}
        \mathbf{z}^\mathit{l} = \sigma(\mathbf{W}_\mathit{z} \ast [\mathbf{f}_\mathit{enc}^\mathit{l}, \hat{\mathbf{s}}^{\mathit{l}+1}]),
    \end{equation}
    \vspace{-2mm}%
    \begin{equation}\label{eqn:lstate}
        \mathbf{s}^\mathit{l} = \mathbf{r}^\mathit{l} \circ \hat{\mathbf{s}}^{\mathit{l}+1},
    \end{equation}
    \vspace{-2mm}%
    \begin{equation}\label{eqn:newinfo}
        \hat{\mathbf{f}}_\mathit{t}^\mathit{l} = \tanh(\mathbf{W}_\mathit{h} \ast [\mathbf{f}_\mathit{enc}^\mathit{l}, \mathbf{s}^\mathit{l}]),
    \end{equation}
    \vspace{-2mm}%
    \begin{equation}\label{eqn:output}
        \mathbf{f}_\mathit{t}^\mathit{l} = (1-\mathbf{z}^\mathit{l}) \circ \hat{\mathbf{s}}^{\mathit{l}+1}
            + \mathbf{z}^{l} \circ \hat{\mathbf{f}}_\mathit{t}^\mathit{l},
    \end{equation}
where $\ast$ denotes the convolution operation, $\circ$ denotes entry-wise product, and $\sigma(\cdot)$ stands for the sigmoid function.

The introduction of the reset gate $\mathbf{r}^\mathit{l}$ and update gate $\mathbf{z}^\mathit{l}$ allows us control the contribution of hidden state, difference attribute vector, and encoder feature in a selective manner. %
Moreover, the convolution transform and linear interpolation in Eqns.~(\ref{eqn:newinfo}) and (\ref{eqn:output}) provide an adaptive means for the transfer of encoder feature and its combination with hidden state.
In comparison to GRU where $\mathbf{f}_\mathit{t}^\mathit{l}$ is adopted as the output of hidden state, we take $\mathbf{s}^\mathit{l}$ as the output of hidden state and $\mathbf{f}_\mathit{t}^\mathit{l}$ as the output of transformed encoder feature.
And experiments empirically validate that such modification can bring moderate gains on attribute generation accuracy.

\subsection{Network Architecture}
Our STGAN is comprised of two components, \ie, a generator $\mathit{G}$ and a discriminator
$\mathit{D}$.
Fig.~\ref{fig:structure} illustrates the network structure of $\mathit{G}$ consisting of an encoder $\mathit{G_{enc}}$ for abstract latent representation and a decoder $\mathit{G_{dec}}$ for target image generation.
The encoder $\mathit{G_{enc}}$ contains five convolution layers with kernel size 4
and stride 2, while the decoder $\mathit{G_{dec}}$ has five transposed convolution layers.
Besides, STU is applied right after each of the first four encoder layers, denoted by $(\mathbf{f}_\mathit{t}^\mathit{l}, \mathbf{s}^\mathit{l}) = \mathit{G_{st}^{\mathit{l}}}(\mathbf{f}_\mathit{enc}^\mathit{l}, \mathbf{s}^{\mathit{l}+1}, \mathbf{att}_\mathit{diff})$.

The discriminator $\mathit{D}$ has two branches $\mathit{D_{adv}}$ and $\mathit{D_{att}}$.
$\mathit{D_{adv}}$ consists of five convolution layers and two fully-connected layers to distinguish whether an image is a fake image or a real one.
$\mathit{D_{att}}$ shares the convolution layers with $\mathit{D_{adv}}$, but predicts an
attribute vector by another two fully-connected layers.
Please refer to the suppl. for more details on the network architecture.

\subsection{Loss Functions}
%
Given an input image $\mathbf{x}$, the encoder features can be obtained by,
    \vspace{-1mm}
    \begin{equation}\label{eqn:G_enc}
        \mathbf{f} = \mathit{G_{enc}}(\mathbf{x}),
    \end{equation}
where $\mathbf{f} = \{\mathbf{f}^{1}_\mathit{enc}, ..., \mathbf{f}^{5}_\mathit{enc}\}$.
Then, guided by $\mathbf{att}_\mathit{diff}$, STUs are deployed to transform encoder features for each layer,
    \vspace{-1mm}
    \begin{equation}\label{eqn:STUnits}
        (\mathbf{f}_\mathit{t}^\mathit{l}, \mathbf{s}^\mathit{l}) = \mathit{G_{st}^\mathit{l}}(\mathbf{f}_\mathit{enc}^\mathit{l}, \mathbf{s}^{\mathit{l}+1}, \mathbf{att}_\mathit{diff}),
    \end{equation}
    %
Note that we adopt four STUs, and directly pass $\mathbf{f}^{5}_\mathit{enc}$ to $\mathit{G_{dec}}$.
The STUs deployed in different layers do not share parameters due to that (i)~the dimensions are different and (ii)~the features of inner layers are more abstract than those of the outer layers.

Let $\mathbf{f}_\mathit{t} = \{\mathbf{f}^{1}_\mathit{t}, ..., \mathbf{f}^{4}_\mathit{t}\}$.
Thus, the editing result of $\mathit{G_{dec}}$ can be given by,
    \vspace{-1mm}
    \begin{equation}\label{eqn:G_dec}
        \hat{\mathbf{y}} = \mathit{G_{dec}}(\mathbf{f}^{5}_\mathit{enc}, \mathbf{f}_\mathit{t}),
    \end{equation}
and can be written by,
    \vspace{-1mm}
    \begin{equation}\label{eqn:G}
        \hat{\mathbf{y}} = G(\mathbf{x}, \mathbf{att}_\mathit{diff}).
    \end{equation}
%
In the following, we detail the reconstruction, adversarial, and attribute manipulation losses which are collaborated to train our STGAN.
\vspace{0.3em}

\noindent\textbf{Reconstruction loss.}
When the target attributes are exactly the same as source ones, \ie, $\mathbf{att}_\mathit{diff} = \mathbf{0}$, it is natural to require that the editing result approximates the source image.
Thus the reconstruction loss is defined as,
    \vspace{-1mm}
    \begin{equation}\label{eqn:rec_loss}
        \mathcal{L}_\mathit{rec} = \|\mathbf{x} - \mathit{G}(\mathbf{x}, \mathbf{0})\|_1,
    \end{equation}
    \vspace{-1mm}
where the $\ell_1$-norm $\| \cdot \|_1$ is adopted for preserving the sharpness of reconstruction result.
\vspace{0.3em}

\noindent\textbf{Adversarial loss.}
When the target attributes are different from source ones, \ie, $\mathbf{att}_\mathit{diff} \neq \mathbf{0}$, the ground-truth of editing result will be unavailable.
Therefore, adversarial loss~\cite{goodfellow2014generative} is employed for constraining the editing result to be indistinguishable from real images.
In particular, we follow Wasserstein GAN (WGAN)~\cite{arjovsky2017wasserstein}
and WGAN-GP~\cite{gulrajani2017improved}, and define the losses for training $D_{adv}$ and $G$ as,
    \vspace{-1mm}
    \begin{equation}\label{eqn:loss_D_adv}
    \begin{split}
        \max\limits_{\mathit{D_{adv}}}\mathcal{L}_\mathit{D_{adv}} =\
            & \mathbb{E}_{\mathbf{x}}\mathit{D_{adv}}(\mathbf{x})
            - \mathbb{E}_{\hat{\mathbf{y}}} \mathit{D_{adv}}(\hat{\mathbf{y}}) +\\
            &\lambda \mathbb{E}_{\hat{\mathbf{x}}} \left[ (\| \nabla_{\hat{\mathbf{x}}} \mathit{D_{adv}}(\hat{\mathbf{x}}) \|_2 - 1)^2 \right],
    \end{split}
    \end{equation}
    \begin{equation}\label{eqn:loss_G_adv}
        \max\limits_\mathit{G}\mathcal{L}_{\mathit{G_{adv}}} =
            \mathbb{E}_{\mathbf{x},\mathbf{att}_\mathit{diff}}
            \mathit{D_{adv}}(G(\mathbf{x},\mathbf{att}_\mathit{diff})),
    \end{equation}%
where $\hat{\mathbf{x}}$ is sampled along lines between pairs of real and generated images.
\vspace{0.3em}

\noindent\textbf{Attribute manipulation loss.}
Even the ground-truth is missing, we can require the editing result to possess the desired target attributes.
Thus, we introduce an attribute classifier $\mathit{D_{att}}$ which shares the convolution layers with $\mathit{D_{adv}}$, and define the following attribute manipulation losses for training $\mathit{D_{att}}$ and generator $\mathit{G}$,
    \vspace{-1mm}
    \begin{equation}\label{eqn:loss_D_att}
    \begin{split}
        \mathcal{L}_\mathit{D_{att}} = -\sum\limits_{\mathit{i}=1}^\mathit{c}
            [&\mathbf{att}_\mathit{s}^{(\mathit{i})}\log{\mathit{D}_\mathit{att}^{(\mathit{i})}(\mathbf{x})} +\\
            &(1-\mathbf{att}_\mathit{s}^{(\mathit{i})})\log{(1-\mathit{D}_\mathit{att}^{(\mathit{i})}(\mathbf{x}))}],
    \end{split}
    \end{equation}
    \vspace{-1mm}
    \begin{equation}\label{eqn:loss_G_att}
    \begin{split}
        \mathcal{L}_\mathit{G_{att}} = -\sum\limits_{\mathit{i}=1}^\mathit{c}
            [&\mathbf{att}_\mathit{t}^{(\mathit{i})}\log{\mathit{D}_\mathit{att}^{(\mathit{i})}(\hat{\mathbf{y}})} +\\
            &(1-\mathbf{att}_\mathit{t}^{(\mathit{i})})\log{(1-\mathit{D}_\mathit{att}^{(\mathit{i})}(\hat{\mathbf{y}}))}],
    \end{split}
    \end{equation}
where $\mathbf{att}_{\mathit{s}/\mathit{t}}^{(\mathit{i})}$ ($\mathit{D}_\mathit{att}^{(\mathit{i})}(\mathbf{x})$) denotes the $\mathit{i}$-th attribute value of $\mathbf{att}_{\mathit{s}/\mathit{t}}$ ($\mathit{D}_\mathit{att}(\mathbf{x})$).
\vspace{0.5em}

\noindent\textbf{Model Objective.}
Taking the above losses into account, the objective to train the discriminator $\mathit{D}$ can be
formulated as,
    \vspace{-1mm}
    \begin{equation}\label{eqn:loss_D}
        \min\limits_\mathit{D}\mathcal{L}_\mathit{D} = -\mathcal{L}_\mathit{D_{adv}} +
            \lambda_{1}\mathcal{L}_\mathit{D_{att}},
    \end{equation}%
and that for the generator $\mathit{G}$ is,
    \vspace{-1mm}
    \begin{equation}\label{eqn:loss_G}
        \min\limits_\mathit{G}\mathcal{L}_\mathit{G} = -\mathcal{L}_\mathit{G_{adv}} +
            \lambda_{2}\mathcal{L}_\mathit{G_{att}} + \lambda_{3}\mathcal{L}_\mathit{rec},
    \end{equation}%
where $\lambda_{1}$, $\lambda_{2}$, and $\lambda_{3}$ are the model tradeoff parameters.

\section{Experiments}\label{section:Experiments}
%
%
We train the model by the ADAM~\cite{kingma2014adam} optimizer with $\beta_1=0.5$ and
$\beta_2=0.999$.
The learning rate is initialized as $2\times10^{-4}$ and decays to $2\times10^{-5}$
for fine-tuning after 100 epochs.
In all experiments, the tradeoff parameters in Eqns.~(\ref{eqn:loss_D}) and~(\ref{eqn:loss_G})
are set to $\lambda_1=1$, $\lambda_2=10$ and $\lambda_3=100$.
All the experiments are conducted in the TensorFlow~\cite{abadi2016tensorflow} environment
with cuDNN 7.1 running on a PC with Intel(R) Xeon(R) E3-1230v5 CPU 3.40GHz and Nvidia GTX1080Ti GPU.
The source code can be found at \url{https://github.com/csmliu/STGAN.git}.

\subsection{Facial Attribute Editing}
Following~\cite{he2017arbitrary, Choi_2018_CVPR}, we first evaluate our STGAN for arbitrary facial attribute editing on the CelebA dataset~\cite{liu2015faceattributes} which has been adopted by most relevant works~\cite{perarnau2016invertible, lample2017fader, he2017arbitrary, Choi_2018_CVPR}.
\vspace{0.5em}

\noindent\textbf{Dataset and preprocessing.}
The CelebA dataset~\cite{liu2015faceattributes} contains 202,599 aligned facial
images cropped to $178\times218$, with 40 \textit{with}/\textit{without} attribute
labels for each image.
The images are divided into training set, validation set and test set.
We take 1,000 images from the validation set to assess the training process, use the rest of the validation set and the training set to train our STGAN model, and utilize the test set for performance evaluation.
We consider 13 attributes, including \textit{Bald,
Bangs, Black Hair, Blond Hair, Brown Hair, Bushy Eyebrows, Eyeglasses, Male, Mouth
Slightly Open, Mustache, No Beard, Pale Skin} and \textit{Young}, due to that
they are more distinctive in appearance and cover most attributes used by the
relevant works.
In our experiment, the central $170\times170$ region of each image is cropped and resized
to $128\times128$ by bicubic interpolation.
Training and inference time please refer to the suppl.
\vspace{0.5em}

\begin{figure*}[t]
    \centering
    \includegraphics[width=\linewidth]{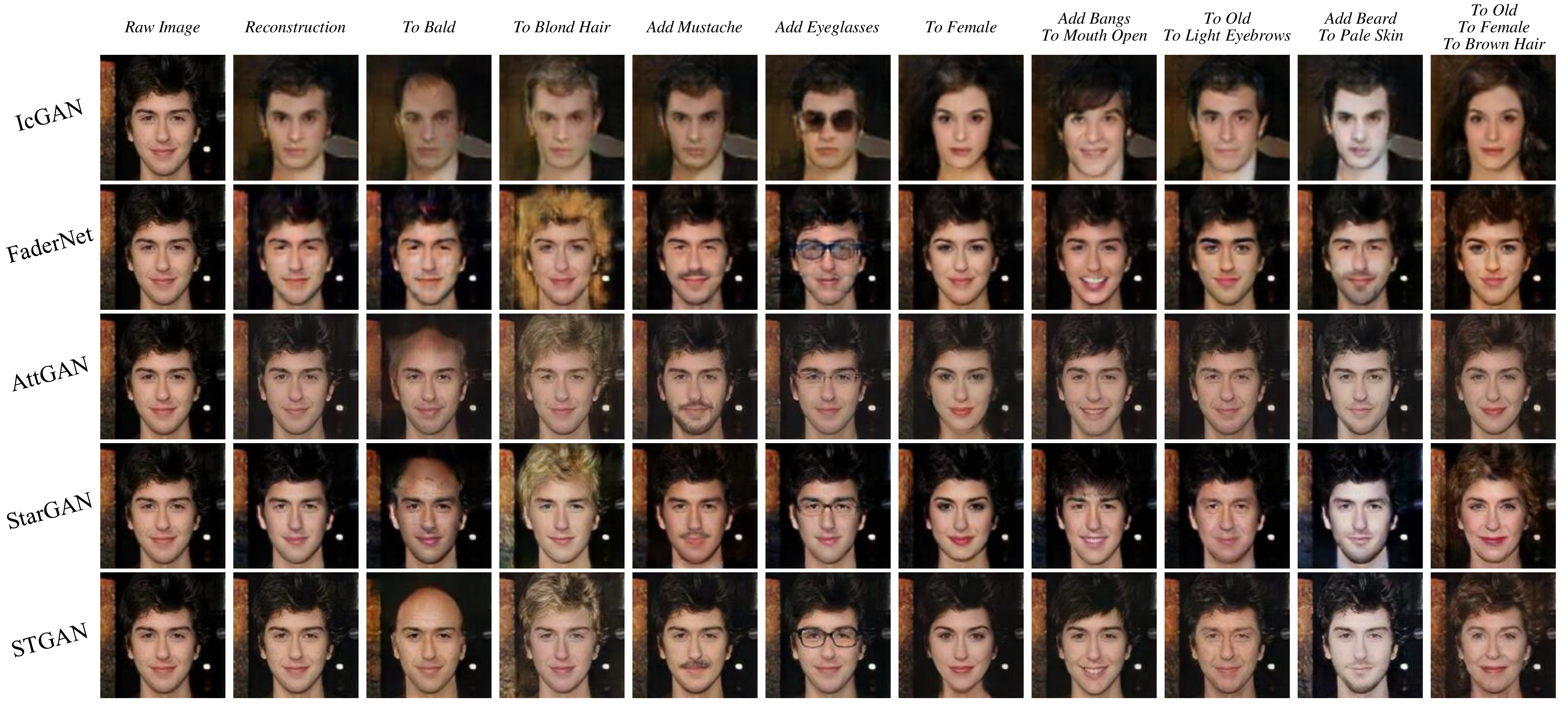}
    \setlength{\abovecaptionskip}{-5pt}
    \setlength{\belowcaptionskip}{0pt}
    \captionsetup{font=small}
    \caption{Facial attribute editing results on the CelebA dataset. The rows from top to down are results of
             IcGAN~\cite{perarnau2016invertible}, FaderNet~\cite{lample2017fader},
             AttGAN~\cite{he2017arbitrary}, StarGAN~\cite{Choi_2018_CVPR} and STGAN.}
    \label{fig:face_image_comp}
    \vspace{-1em}
\end{figure*}

\begin{table}
    \centering
    \scalebox{0.65}{
    \begin{tabular}{lccccc}
        \hline
        {Method} & IcGAN & FaderNet & AttGAN & StarGAN & STGAN \\ \hline
        {PSNR/SSIM} & 15.28/0.430 & 30.62/0.908 & 24.07/0.841 & 22.80/0.819 & \textbf{31.67}/\textbf{0.948} \\ \hline
    \end{tabular} }
    \setlength{\abovecaptionskip}{3pt}
    \setlength{\belowcaptionskip}{0pt}
    \captionsetup{font=small}
    \caption{Reconstruction quality of the comparison methods on facial attribute editing task.}
    \label{tab:face_rec_comp}
\end{table}

\begin{table}[t]
    \centering
    \scalebox{.7}{
    \begin{tabular}{lcccccc}
        \hline
        {Method} & \textit{Bald} & \textit{Bangs} & \textit{Eyebrows} & \textit{Glasses}
                        & \textit{\tabincell{c}{Hair\\Color}} & \textit{Male} \\ \hline
        AttGAN & 12.76\% & 34.28\% & 10.64\% & 30.04\% & 11.52\% & 15.68\% \\
        StarGAN & 11.28\% & 18.12\% & 19.40\% & 19.20\% & 32.28\% & 13.52\% \\
        STGAN & \textbf{75.96\%} & \textbf{47.60\%} & \textbf{69.96\%} & \textbf{50.76\%}
              & \textbf{56.20\%} & \textbf{70.80\%} \\ \hline \hline
        {Method} & \textit{\tabincell{c}{Mouth\\Open}} & \textit{Mustache} & \textit{\tabincell{c}{No\\Beard}}
                        & \textit{\tabincell{c}{Pale\\Skin}} & \textit{Young} & \textit{Average} \\ \hline
        AttGAN & 20.40\% & 20.20\% & 18.92\% & 21.08\% & 15.16\% & 19.15\% \\
        StarGAN & 23.40\% & 10.04\% & 20.36\% & 16.52\% & 27.92\% & 19.27\% \\
        STGAN & \textbf{56.20\%} & \textbf{69.76\%} & \textbf{60.72\%} & \textbf{62.40\%}
              & \textbf{56.92\%} & \textbf{61.58\%} \\ \hline
    \end{tabular}
    }
    \setlength{\abovecaptionskip}{5pt}
    \setlength{\belowcaptionskip}{0pt}
    \captionsetup{font=small}
    \caption{Results of user study for ranking the models on facial attribute
    editing task.}
    \label{tab:face_user_study}
\end{table}

\begin{table}
    \centering
    \scalebox{.8}{
    \begin{tabular}{lcccc}
        \hline
        {Method} & AttGAN & StarGAN & CycleGAN & STGAN \\ \hline
        \textit{summer}$\rightarrow$\textit{winter} & 4.7\% & 9.9\% & 24.9\% & \textbf{60.5\%} \\
        \textit{winter}$\rightarrow$\textit{summer} & 17.0\% & 7.9\% & 24.6\% & \textbf{50.5\%} \\ \hline
    \end{tabular}
    }
    \setlength{\abovecaptionskip}{0pt}
    \setlength{\belowcaptionskip}{-.5em}
    \captionsetup{font=small}
    \caption{Results of user study for ranking the models on season conversion task.}
    \label{tab:season_user_study}
\end{table}

\noindent\textbf{Qualitative results.}
We compare STGAN with four competing methods, \ie, IcGAN~\cite{perarnau2016invertible}, FaderNet~\cite{lample2017fader}, AttGAN~\cite{he2017arbitrary} and StarGAN~\cite{Choi_2018_CVPR}.
The qualitative results are shown in Fig.~\ref{fig:face_image_comp}.
%
%
The results of AttGAN are generated by the released model,
and we retrain other models for a fair comparison.
It can be observed from Fig.~\ref{fig:face_image_comp}, all the competing methods are still limited in manipulating complex attributes, \eg, \emph{Bald}, \emph{Hair}, and \emph{Age}
%
%
%
, and are prone to over-smoothing results.
Besides, their results are more likely to be insufficiently modified and photo non-realistic when dealing with complex and/or multiple attributes.
In comparison, our STGAN is effective in correctly manipulating the desired attributes, and can produce results with high image quality.
More editing results are given in the suppl.
\vspace{0.5em}

\begin{figure}[t]
    \centering
    \includegraphics[width=\linewidth]{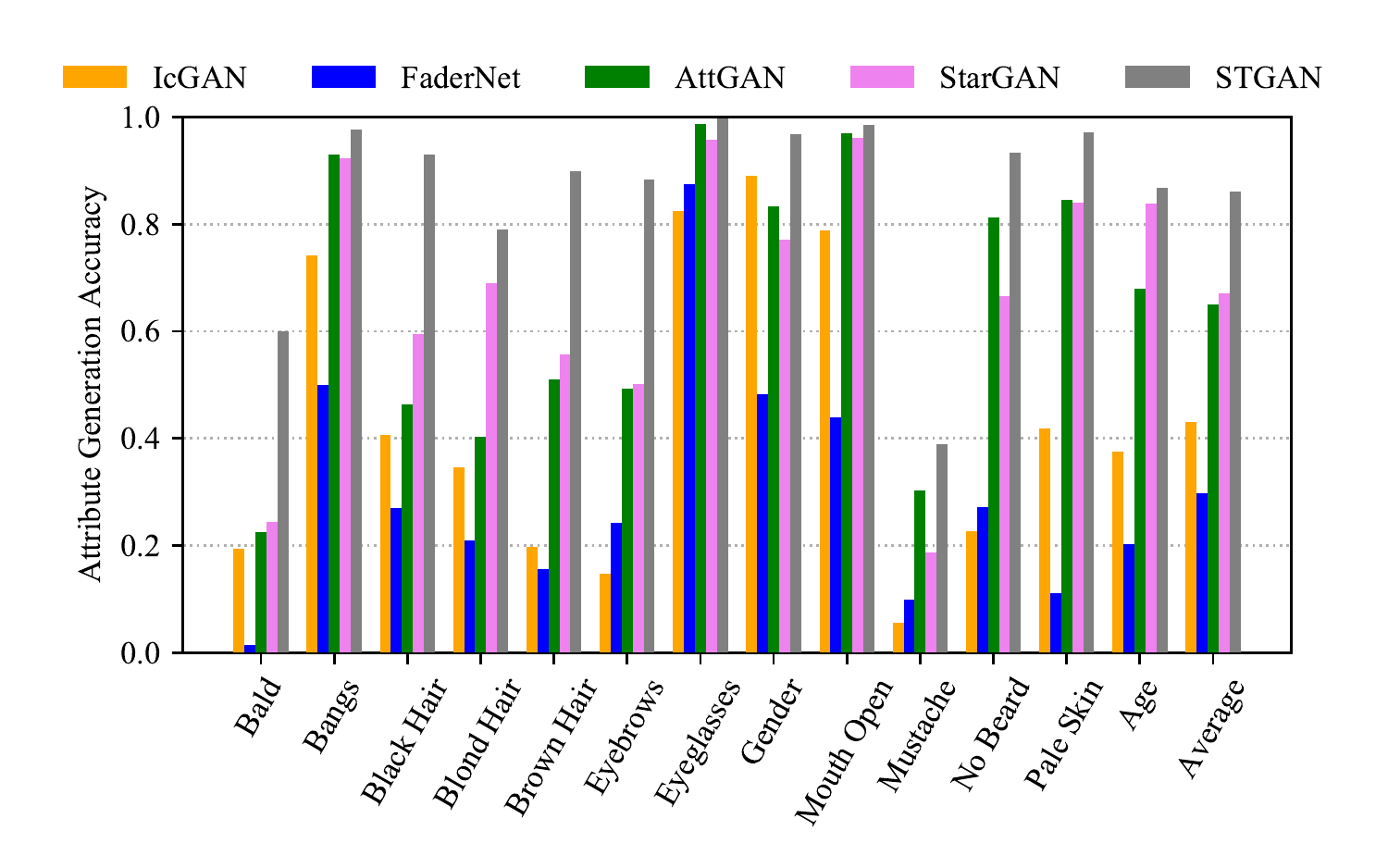}
    \setlength{\abovecaptionskip}{-10pt}
    \setlength{\belowcaptionskip}{0pt}
    \captionsetup{font=small}
    \caption{Attribute generation accuracy of IcGAN~\cite{perarnau2016invertible}, FaderNet~\cite{lample2017fader},
             AttGAN~\cite{he2017arbitrary}, StarGAN~\cite{Choi_2018_CVPR} and STGAN.}
    \label{fig:face_acc_comp}
    \vspace{-1.5em}
\end{figure}

\noindent\textbf{Quantitative evaluation.}
The performance of attribute editing can be evaluated from two aspects, \ie, image quality and attribute generation accuracy.
Due to the unavailability of editing result, we resort to two alternative measures for quantitative evaluation of our STGAN.
First, we use the training set of STGAN to train a deep attribute classification model which can attain an accuracy of 94.5\% for the 13 attributes on the test set.
Then Fig.~\ref{fig:face_acc_comp} shows the attribute generation accuracy, \ie, classification accuracy on the changed attributes of editing results.
It can be seen that our STGAN outperforms all the competing methods with a large margin.
For the attributes \emph{Bald}, \emph{Black Hair}, \emph{Brown Hair}, and \emph{Eyebrows}, STGAN achieves 20\% accuracy gains against the competing methods.

As for image quality, we keep target attribute vector the same as the source one, and give the the PSNR/SSIM results of reconstruction in Table \ref{tab:face_rec_comp}.
Benefited from the STUs and difference attribute vector, our STGAN achieves much better reconstruction ($> 7$ dB by PSNR) in comparison to AttGAN and STGAN.
The result is consistent with Fig. \ref{fig:face_rec_image}.
%
The reconstruction ability of IcGAN is very limited due to the training procedure.
FaderNet obtains better reconstruction results, mainly ascribing to that each FaderNet model is
trained to deal with only one attribute.%
\vspace{0.5em}

\noindent\textbf{User study.}
User study on a crowdsourcing platform is conducted to evaluate the generation quality of three top-performance methods, \ie, AttGAN, StarGAN and STGAN.
We consider 11 tasks for 13 attributes, as the transfer among \textit{Blond
Hair, Black Hair} and \textit{Brown Hair} are merged into \textit{Hair Color}.
For each task, 50 validated people participate in and each of them is given 50
questions.
In each question, people are given a source image randomly selected from test set and
the editing results by AttGAN, StarGAN and STGAN.
For a fair comparison, the results are shown in a random order.
The users are instructed to choose the best result which \emph{changes the
attribute more successfully}, \emph{is of higher image quality} and \emph{better preserves
the identity and fine details of source image}.
The results are shown in Table~\ref{tab:face_user_study}, and STGAN has higher probability to be selected as the best method on all the 11 tasks.
%

\subsection{Season Translation}
We further train our STGAN for image-to-image translation between \textit{summer} and
\textit{winter} using the dataset released by CycleGAN~\cite{zhu2017unpaired}.
The dataset contains photos of Yosemite, including 1,231 summer and 962 winter images in the training set, and 309 summer and 238 winter images for testing.
We also randomly select 100 images from the training set to validation.
All images are used as the original size of $256\times256$.

\begin{figure}[t]
    \centering
    \captionsetup[subfigure]{labelformat=simple, listofformat=subsimple}
    \begin{minipage}{0.18\linewidth}
        \includegraphics[width=\linewidth]{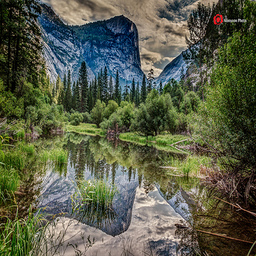}
        \includegraphics[width=\linewidth]{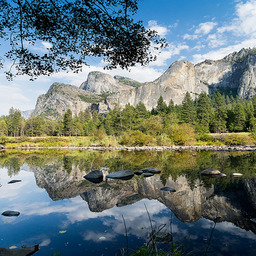}
        \includegraphics[width=\linewidth]{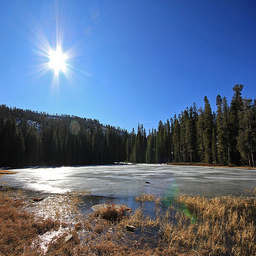}
        \includegraphics[width=\linewidth]{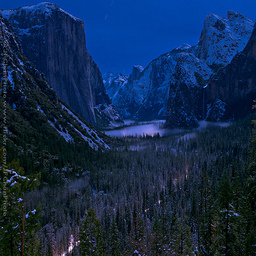}
        \vspace{-2em}
        \caption*{\scriptsize Input}
    \end{minipage}
    \begin{minipage}{0.18\linewidth}
        \includegraphics[width=\linewidth]{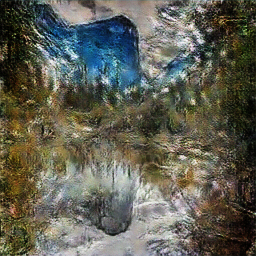}
        \includegraphics[width=\linewidth]{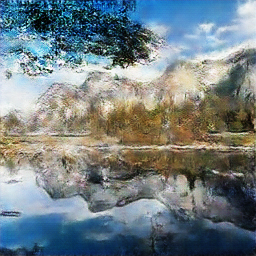}
        \includegraphics[width=\linewidth]{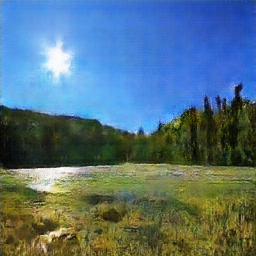}
        \includegraphics[width=\linewidth]{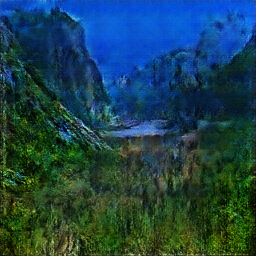}
        \vspace{-2em}
        \caption*{\scriptsize AttGAN}
    \end{minipage}
    \begin{minipage}{0.18\linewidth}
        \includegraphics[width=\linewidth]{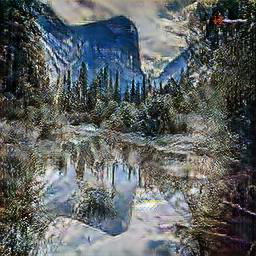}
        \includegraphics[width=\linewidth]{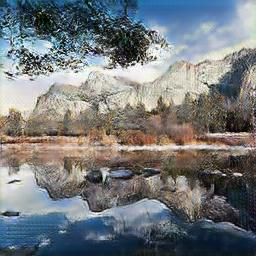}
        \includegraphics[width=\linewidth]{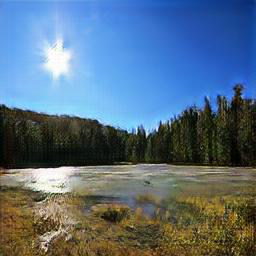}
        \includegraphics[width=\linewidth]{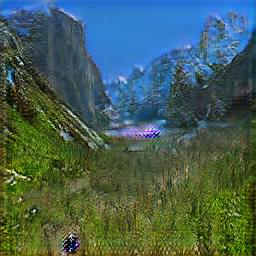}
        \vspace{-2em}
        \caption*{\scriptsize StarGAN}
    \end{minipage}
    \begin{minipage}{0.18\linewidth}
        \includegraphics[width=\linewidth]{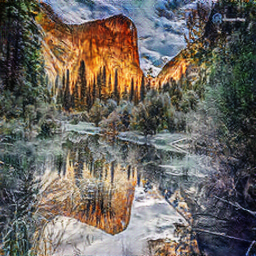}
        \includegraphics[width=\linewidth]{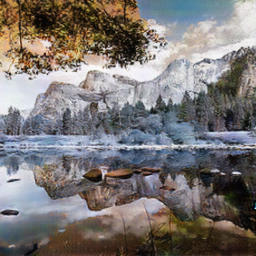}
        \includegraphics[width=\linewidth]{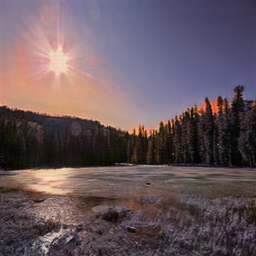}
        \includegraphics[width=\linewidth]{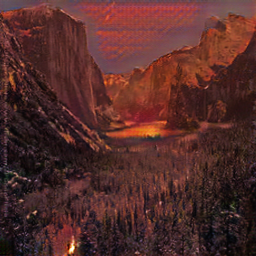}
        \vspace{-2em}
        \caption*{\scriptsize CycleGAN}
    \end{minipage}
    \begin{minipage}{0.18\linewidth}
        \includegraphics[width=\linewidth]{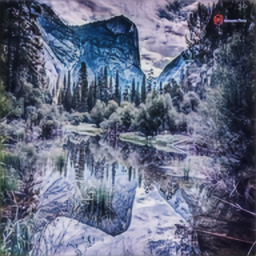}
        \includegraphics[width=\linewidth]{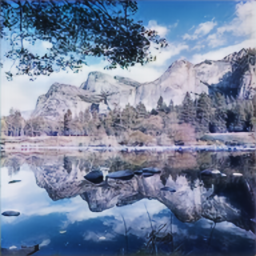}
        \includegraphics[width=\linewidth]{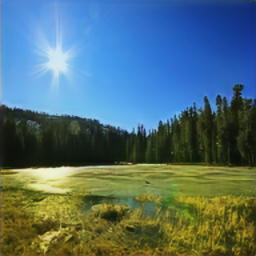}
        \includegraphics[width=\linewidth]{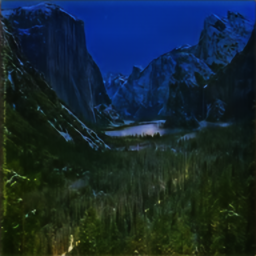}
        \vspace{-2em}
        \caption*{\scriptsize STGAN}
    \end{minipage}
    \setlength{\abovecaptionskip}{5pt}
    \setlength{\belowcaptionskip}{0pt}
    \captionsetup{font=small}
    \caption{Results of season translation, the top two rows are \textit{summer}$\rightarrow$\textit{winter},
    and the bottom two rows are \textit{winter}$\rightarrow$\textit{summer}.}
    \label{fig:season_image_comp}
    \vspace{-0.75em}
\end{figure}

%
We compare our STGAN with AttGAN~\cite{he2017arbitrary}, StarGAN~\cite{Choi_2018_CVPR}, and CycleGAN released by Zhu \etal~\cite{zhu2017unpaired}.
Note that CycleGAN uses two generators respectively for \textit{summer}$\rightarrow$\textit{winter}
and \textit{winter}$\rightarrow$\textit{summer} translation, while the other three methods conduct the two tasks with a single model.
Fig.~\ref{fig:season_image_comp} shows several examples of translation results.
It can be seen that STGAN performs favorably against the competing methods.
We also conduct a user study using the same setting for facial attribute editing.
From Table~\ref{tab:season_user_study}, our STGAN has a probability of more than 50\% to win among the four competing methods.

\begin{figure}[t]
    \centering
    \includegraphics[width=\linewidth]{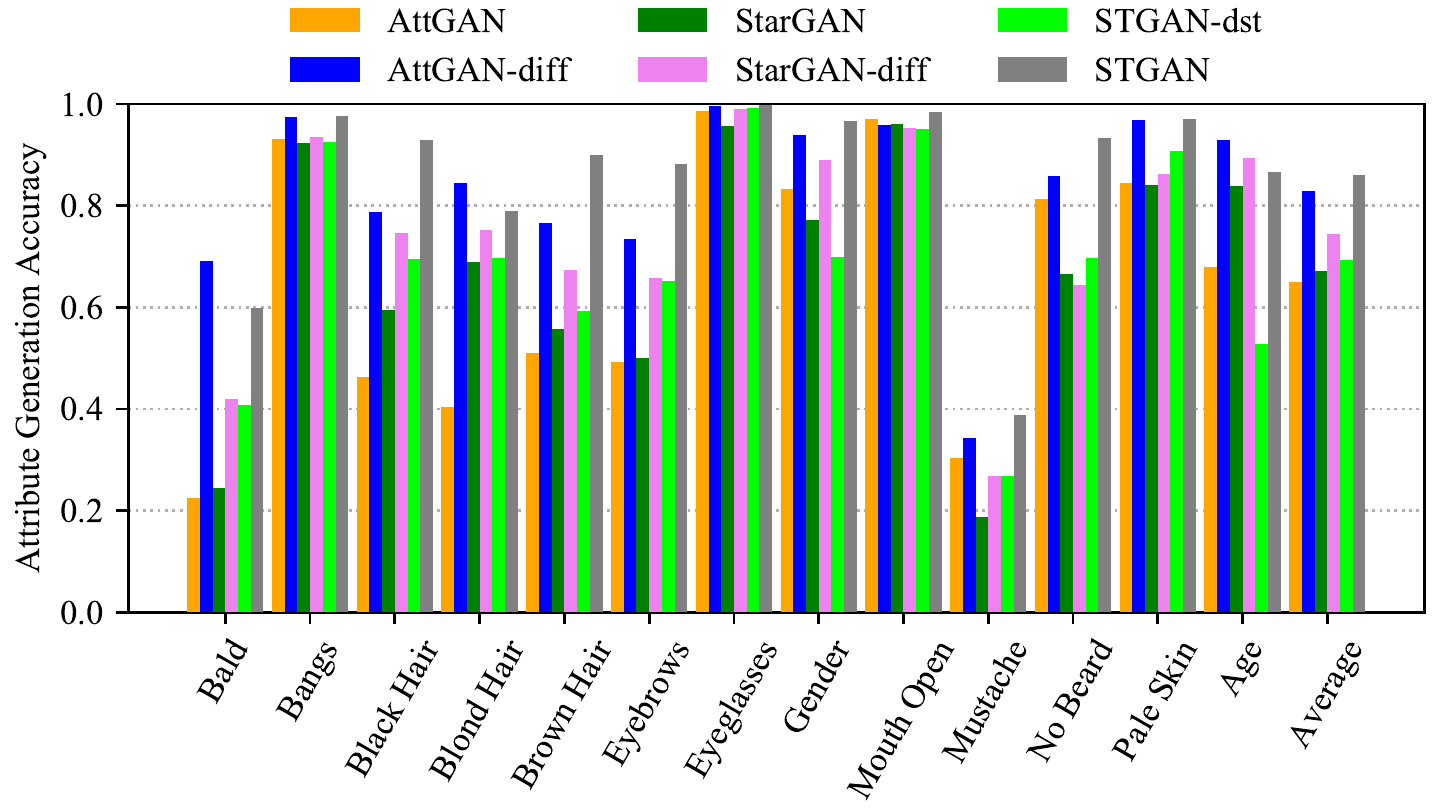}
    \setlength{\abovecaptionskip}{-8pt}
    \setlength{\belowcaptionskip}{0pt}
    \captionsetup{font=small}
    \caption{Effect of difference attribute vector on AttGAN, StarGAN and STGAN.}
    \label{fig:diff_label_acc}
    \vspace{-0.75em}
\end{figure}

\begin{figure}[t]
    \centering
    \includegraphics[width=\linewidth]{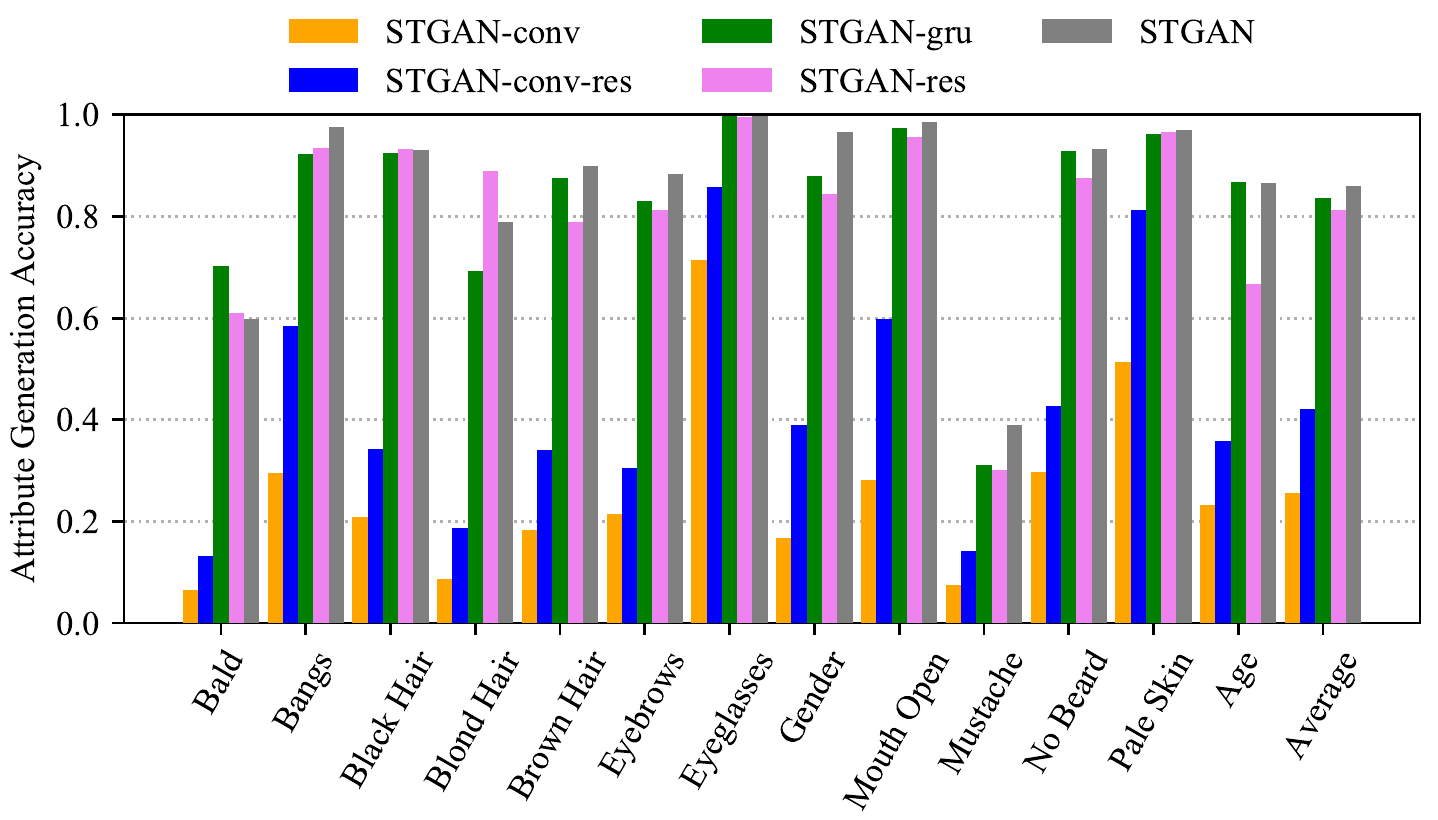}
    \setlength{\abovecaptionskip}{-8pt}
    \setlength{\belowcaptionskip}{0pt}
    \captionsetup{font=small}
    \caption{Attribute generation accuracy of STGAN variants.}
    \label{fig:ablation_acc}
    \vspace{-1.25em}
\end{figure}

\vspace{-0.5em}
\section{Ablation Study}\label{section:AblationStudy}
\vspace{-0.5em}

Using facial attribute editing, we implement several variants of STGAN, and evaluate them on CelebA~\cite{liu2015faceattributes} to assess the role of difference attribute vector and STUs.
Concretely, we consider six variants, \ie,
(i) STGAN: original STGAN,
(ii) STGAN-dst: substituting difference attribute vector with target attribute vector,
(iii) STGAN-conv: instead of STU, applying a convolution operator by taking encoder feature and difference attribute vector as input to modify encoder feature,
(iv) STGAN-conv-res: adopting the residual learning formulation to learn the convolution operator in STGAN-conv,
(v) STGAN-gru: replacing STU with GRU in STGAN,
(vi) STGAN-res: adopting the residual learning formulation to learn the STU in STGAN.
We also train AttGAN and StarGAN models with difference attribute vector, denoted by AttGAN-diff and StarGAN-diff.
Figs.~\ref{fig:diff_label_acc} and~\ref{fig:ablation_acc} show their results on attribute manipulation. Please refer to the suppl. for qualitative results.

\vspace{0.5em}

\noindent\textbf{Difference attribute vector \vs target attribute vector.}
In Fig.~\ref{fig:diff_label_acc}, we present the comparison results of AttGAN, StarGAN and STGAN-dst with their counterparts (\ie, AttGAN-diff, StarGAN-diff and STGAN) by using difference attribute vector.
One can see that difference attribute vector generally benefit attribute generation accuracy for all the three models.
Moreover, empirical studies show that the use of difference attribute vector gives rise to training stability as well as image reconstruction performance.
Note that while AttGAN-diff and StarGAN-diff perform better than AttGAN and StarGAN, they still suffer from the poor image quality.
\vspace{0.5em}

\noindent\textbf{Selective Transfer Unit \vs its variants.}\label{subsection:ablation_STU}
%
Fig.~\ref{fig:ablation_acc} reports the attribute generation accuracy of several STGAN variants for transforming encoder feature conditioned on difference attribute vector.
The two convolutional methods, \ie, STGAN-conv and STGAN-conv-res, are significantly inferior to STGAN, indicating that they are limited in selective transfer of encoder feature.
In comparison to STGAN-conv, STGAN-conv-res achieves relatively higher attribute generation accuracy.
So we also compare STGAN with STGAN-res to check whether STU can be improved via residual learning.
However, due to the selective ability of STUs, further deployment of residual learning cannot bring any gains for most attributes, and performs worse for several global (\eg, \emph{Gender}, \emph{Age}) and fine (\eg, \emph{Mustache}, \emph{Beard}) attributes.
Finally, STGAN is compared with STGAN-gru by using transformed feature as hidden state.
Although STGAN-gru performs better on \emph{Bald}, STGAN is slightly superior to STGAN-gru for most attributes and the gain is notable for attributes \emph{Gender} and \emph{Mustache}.

\vspace{-0.5em}
\section{Conclusion}\label{section:Conclution}
%
\vspace{-0.5em}
In this paper, we study the problem of arbitrary image attribute editing for selective transfer perspective, and present a STGAN model by incorporating difference attribute vector and selective transfer units (STUs) in encoder-decoder network.
By taking difference attribute vector rather than target attribute vector as model input, our STGAN can focus on editing the attributes to be changed, which greatly improves the image reconstruction quality and enhances the flexible translation of attributes.
Furthermore, STUs are presented to adaptively select and modify encoder feature tailored to specific attribute editing task, thereby improving attribute manipulation ability and image quality simultaneously.
Experiments on arbitrary facial attribute editing and season translation show that our STGAN performs favorably against state-of-the-arts in terms of attribute generation accuracy and image quality of editing results.

\vspace{0.5em}
\noindent\textbf{Acknowledgement.} This work was supported in part by the National Natural Science Foundation of China under grant No. 61671182 and 61872118.

{\small
\bibliographystyle{ieee}
\bibliography{PaperID_1439}
}

\end{document}